\documentclass[manuscript,screen,nonacm]{acmart}

\usepackage{bm}
\usepackage{nicefrac}
\usepackage{enumitem}

\usepackage[capitalize,noabbrev]{cleveref}
\crefname{appendix}{appendix}{appendices}
\Crefname{appendix}{Appendix}{Appendices}

\def\vx{{\bm{x}}}
\def\vy{{\bm{y}}}
\def\vk{{\bm{k}}}

\newcommand{\E}{\mathbb{E}}
\newcommand{\R}{\mathbb{R}}

\DeclareMathOperator*{\argmax}{arg\,max}

\newtheorem{remark}{Remark}[section]
\theoremstyle{definition}
\newtheorem{definition}{Definition}[section]

\begin{document}

\title{Boundary Variance Inflation Causes Acquisition Bias in Gaussian Processes}

\author{Maria B\aa{}nkestad}
\authornote{Corresponding author, maria.bankestad@ri.se.}
\email{maria.bankestad@ri.se}
\affiliation{%
  \institution{RISE Research Institutes of Sweden}
  \city{Stockholm}
  \country{Sweden}
}

\author{Sanna Jarl}
\affiliation{%
  \institution{RISE Research Institutes of Sweden}
  \city{Stockholm}
  \country{Sweden}
}

\author{Jens Sj\"olund}
\affiliation{%
  \institution{Uppsala University}
  \city{Uppsala}
  \country{Sweden}
}

\begin{abstract}
Gaussian processes with stationary kernels on bounded domains exhibit inflated posterior variance near the boundary. Despite being a long-recognized artifact in geostatistics and a source of over-exploration in Bayesian optimization, the causes and effects of boundary-induced acquisition bias are underexplored.
We trace the root cause to a simple geometric mechanism: the truncation of the kernel correlation neighborhood at the domain boundary creates an observation-independent distortion that worsens with dimensionality.
We show how this distortion manifests across three acquisition classes: variance maximization concentrates selections at the corners, whereas negative integrated posterior variance and expected predictive information gain move selections inward to axis-aligned interior shells.
These patterns arise without reference to any objective function, meaning that acquisition behavior can be dominated by kernel geometry rather than the desired task-specific uncertainty.
To quantify this, we introduce a function-free selection-profile diagnostic for arbitrary acquisitions, kernels, and bounded-domain geometries.\looseness-1
\end{abstract}

\keywords{Gaussian processes, posterior variance, boundary bias, Bayesian optimization, experimental design, acquisition functions, active learning}

\maketitle

\section{Introduction}
\label{sec:introduction}

Gaussian processes (GPs) are widely used as surrogate models in Bayesian optimization (BO)~\cite{shahriari2016taking, garnett2023bayesian} and Bayesian experimental design (BED)~\cite{mackay1992information, chaloner1995bayesian, rainforth2024modern}.
In both settings, the next observation is selected by optimizing acquisition functions that rely, directly or indirectly, on the GP posterior variance.
On bounded domains such as $[0,1]^D$, this variance is susceptible to a geometric artifact where the posterior variance is artificially inflated at the boundaries, regardless of the function values.

Such boundary effects are documented in the kriging literature~\cite{cressie1993statistics}.
In GP-based sensor placement, \citet{krause2008near} noted that entropy-based criteria tend to cluster sensors at the boundary, ``wasting'' sensed information, and proposed mutual information as a remedy.
\citet{siivola2018boundary} identified analogous boundary over-exploration in BO and proposed virtual derivative sign observations as a per-acquisition correction.
Boundary-aware GP constructions~\cite{solin2019boundary, gulian2022boundary} encode physical boundary conditions when known, while normalized convolution~\cite{knutsson1993normalized} addresses the analogous loss of support in image filtering by reweighting and renormalizing convolution with a certainty field.

Variance-based design criteria, such as IMSPE~\cite{gramacy2020surrogates}, mitigate certain symptoms through global integration, but the connection to the boundary mechanism is rarely made explicit.
Despite this prior work, boundary-induced acquisition bias is seldom isolated in the development or benchmarking of acquisition functions.
The geometric mechanism is rarely separated from objective-function behavior, the differential effect across acquisition families is not characterized, and the spatial selection patterns are not routinely measured.

Our contribution is to isolate this variance-inflation mechanism and characterize how a single geometric effect propagates across three variance-driven acquisition classes.
In short, the bias stems from the kernel correlation neighborhood being truncated at the domain boundary, causing boundary points to have less in-domain correlated volume than comparable interior points.
The effect is observation-independent: for fixed hyperparameters, it depends on the kernel, bounded domain, and the training locations, but \emph{not} on the observed function values.
One-step variance maximization (VM), which includes the Gaussian one-point expected information gain (EIG)~\cite{lindley1956eig} and the upper confidence bound (UCB) exploration term~\cite{srinivas2010gaussian}, concentrates selections at corners.
Negative integrated posterior variance (NIPV)~\cite{seo2000nipv,cohn1996active} and expected predictive information gain (EPIG)~\cite{bickfordsmith2023prediction} instead select axis-aligned interior shells, with NIPV closer to the boundary than EPIG.
These patterns arise without reference to any particular objective function, showing that acquisition behavior can be dominated by kernel geometry rather than task-specific uncertainty.

We quantify this effect with a function-free \emph{selection-profile diagnostic}, applicable to arbitrary acquisitions, kernels, and domain geometries, and validate it against sequential acquisition on representative test functions.

\section{Background}
\label{sec:background}

We consider an unknown function $f\colon [0,1]^D \to \R$ modeled by a GP~\cite{rasmussen2006gaussian},
$z \sim \mathcal{GP}(m(\vx), k_\theta(\vx, \vx'))$, with noisy observations
$y_i = z(\vx_i) + \varepsilon_i$, $\varepsilon_i \sim \mathcal{N}(0, \sigma_n^2)$.
The posterior variance at a test point $\vx_*$ and the posterior covariance between any two points are
\begin{equation}
s^2(\vx_*) = k_\theta(\vx_*, \vx_*) - \vk_*^\top (K + \sigma_n^2 I)^{-1} \vk_*, \qquad
c(\vx, \vx') = k_\theta(\vx, \vx') - \vk_\vx^\top (K + \sigma_n^2 I)^{-1} \vk_{\vx'},
\label{eq:gp_var_cov}
\end{equation}
where $K_{ij} = k_\theta(\vx_i, \vx_j)$, $\vk_\vx = (k_\theta(\vx_1, \vx), \ldots, k_\theta(\vx_N, \vx))^\top$, and the variance is the diagonal case $s^2(\vx) = c(\vx, \vx)$.
\begin{remark}[Observation-independence of the variance field]
\label{rem:y_independence}
Both $s^2$ and $c$ depend on the training locations $X$ and the kernel $k_\theta$, but not on the observed values $\vy$, which only enter the posterior through the mean.
Any acquisition function that is a functional of $s^2$ and $c$ alone therefore inherits this observation-independence: its spatial preferences are determined by the kernel geometry and the training locations alone.
\end{remark}

The class of acquisition functions whose spatial preferences are determined by $s^2$ and $c$ (directly or through posterior correlations) is large, encompassing several of the most common choices.
For target points $\{\vx_j^*\}_{j=1}^M$, let $c_j = c(\vx, \vx_j^*)$ denote the posterior covariance between a candidate $\vx$ and target $\vx_j^*$.
The corresponding squared posterior correlation is
\[
    \rho_j^2 =
    \frac{c_j^2}{(s^2(\vx) + \sigma_n^2)(s^2(\vx_j^*) + \sigma_n^2)} .
\]
The three acquisition classes we consider, variance maximization, negative integrated posterior variance, and expected predictive information gain, are
\begin{equation}
    \mathrm{VM}(\vx) = s^2(\vx), \qquad
    \mathrm{NIPV}(\vx) = \textstyle\sum_j \frac{c_j^2}{s^2(\vx) + \sigma_n^2}, \qquad
    \mathrm{EPIG}(\vx) = \textstyle\frac{1}{M}\sum_j h(\rho_j^2),
\end{equation}
with $h(\rho^2) = -\tfrac{1}{2}\log(1-\rho^2)$.
We view variance maximization as an umbrella label that includes the GP-UCB exploration term and GP EIG,
$\tfrac{1}{2}\log\bigl(1 + s^2(\vx)/\sigma_n^2\bigr)$, since those are monotone transforms of $s^2$ and therefore share the same argmax as pure variance maximization~\cite{mackay1992information}.

NIPV and EPIG both use the posterior cross-covariances $c_j$ between candidates and targets, but aggregate them differently.
NIPV measures the total posterior variance removed by observing at $\vx$, whereas EPIG averages pairwise mutual information between candidate and target points through the convex function $h$.
We analyze EPIG in the prediction-oriented setting where targets span the full domain, so all three methods share the same candidate space and can be compared directly.
In BO, full UCB $\mu(\vx) + \beta\sigma(\vx)$ depends on $\vy$ through $\mu$, but its exploration component does not.
When exploration dominates early in optimization, this variance bias controls where observations are placed.\looseness-1

\section{Boundary Variance Inflation}
\label{sec:boundary_bias}

The posterior variance $s^2(\vx_*)$ is the prior variance minus a reduction determined by the covariance between $\vx_*$ and the training inputs.
For stationary kernels, this covariance decays with distance at a rate determined by the lengthscale $\ell$.
On a bounded domain, the correlation neighborhood of a boundary point is truncated: training data can contribute from the interior side but not from beyond the boundary.
\begin{definition}[Boundary distance]
\label{def:boundary_dist}
For $\vx \in [0,1]^D$, the boundary distance is $d_\partial(\vx) = \min_{d} \min(x_d, 1 - x_d)$, the minimum distance from $\vx$ to any face of the unit hypercube. This ranges from $0$ on any face to $1/2$ at the cube's center.
\end{definition}

An interior point with $d_\partial(\vx) \gg \ell$ has an approximately complete correlation neighborhood within the domain; a boundary point with $d_\partial(\vx) \ll \ell$ has this neighborhood truncated by one or more faces.
On average, the boundary point has higher posterior variance than an interior point with comparable local training density.
The effect compounds with dimension: the boundary truncates the correlation neighborhood independently in each coordinate, so corner points lose correlated volume along all $D$ faces simultaneously.
For typical $N$ and $D$, this systematically elevates posterior variance at corners and edges relative to the interior.
Because this inflated variance enters the acquisition functions through different functional expressions, the resulting artifacts differ.

\begin{figure}[t]
    \centering
    \includegraphics[width=0.97\textwidth]{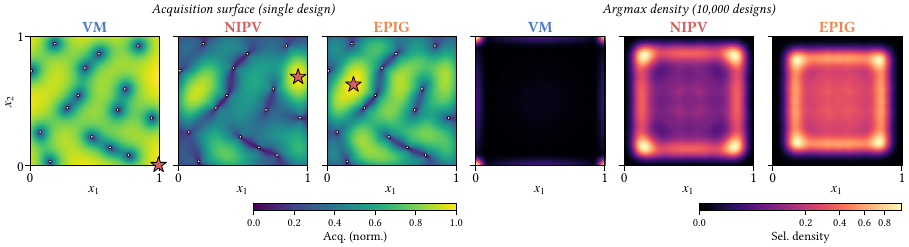}
      \caption{
          Acquisition surfaces and argmax-location densities on $[0,1]^2$ under an isotropic Mat\'ern-$\nicefrac{5}{2}$ kernel ($\ell =
       0.2$, $N = 15$).
          \textbf{Left:} Acquisition value, individually normalized, for one fixed Sobol training design; $\bigstar$ marks the argmax.
          \textbf{Right:} D4-symmetrized argmax-location density over $10\,000$ random training sets.
          VM concentrates at the corners, NIPV forms a near-boundary square frame, and EPIG forms a frame further inside.
          These are 2D instances of the axis-aligned interior shells discussed throughout, determined by kernel geometry and training locations rather than observed function values.}\looseness-1
    \Description{Two-by-three grid of acquisition surfaces (left) and argmax densities (right) on the unit square for VM, NIPV, and EPIG. VM peaks at corners, NIPV along edges away from corners, EPIG in the interior.}
    \label{fig:surface_2d}
\end{figure}

VM inherits the artifact directly, concentrating observations at the boundary because the acquisition is monotone in $s^2(\vx)$.
NIPV partially corrects it: the high local variance $s^2(\vx)$ at a boundary candidate inflates the denominator, while the smaller cross-covariances $c_j$ with the target set depress the numerator.
The maximum sits on an axis-aligned interior shell just inside the boundary.
EPIG shifts selection further inward through the soft-threshold structure of $h$, which is steep near $\rho^2 = 1$ and shallow near zero: each target contributes substantially only if the candidate has high posterior correlation with it.
An interior candidate achieves this with targets in all directions; a boundary candidate's truncated correlations mean that most distant targets contribute little to the average.
Consequently, EPIG's maximum sits on a shell at a larger $d_\partial$ than NIPV's.

Under fixed hyperparameters, none of these spatial preferences responds to the observed function values (\Cref{rem:y_independence}); in particular, none can, by itself, identify regions where the realized function varies most.
The boundary bias is therefore a property of the variance field itself; the acquisition function determines only how that bias is translated into a placement preference.

\Cref{fig:surface_2d} illustrates the effect in two dimensions.
The left panels show one representative acquisition surface for each method, computed for a fixed Sobol design under an isotropic Mat\'ern-$\nicefrac{5}{2}$ kernel ($\ell = 0.2$, $N = 15$).
VM peaks sharply at a domain corner ($\bigstar$); NIPV peaks near the boundary but away from the corner; EPIG peaks in the interior.
The right panels show the argmax-location density over 10\,000 random training designs.
Together, these show the bias reflects kernel geometry and training locations, not observed function values.

\section{The Selection Profile Diagnostic}
\label{sec:diagnostic}
The diagnostic tracks where the argmax occurs, rather than the mean acquisition value at each boundary distance.
This matters because sequential design selects maximizers, and mean profiles can be nearly flat even when the argmax repeatedly falls in specific boundary-distance shells.
We define the \emph{selection profile} as the empirical distribution of $d_\partial(\vx_\mathrm{next})$ over many independent training sets.
This gives a one-dimensional summary in any dimension.

We run a \emph{controlled diagnostic}: for fixed $D$, $N$, and kernel hyperparameters, we draw $S = 1000$ random training sets on $[0,1]^D$, compute the GP posterior under a Mat\'ern-$\nicefrac{5}{2}$ kernel, and record the boundary-distance bin of the acquisition argmax.
To separate acquisition preference from cube geometry, candidates are stratified across boundary-distance shells and then reweighted to recover the uniform-sampling distribution.

\begin{figure}[h]
    \centering
    \includegraphics[width=0.95\textwidth]{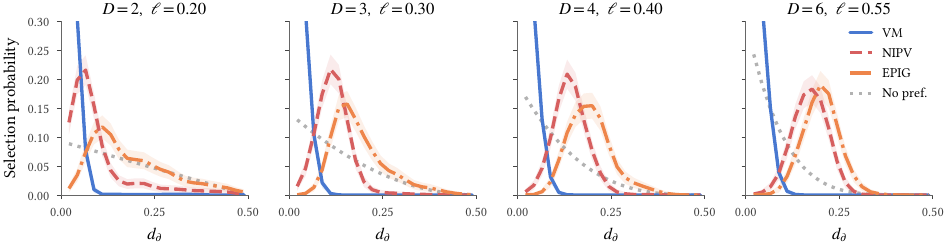}
    \caption{%
        Selection profiles under a Mat\'ern-$\nicefrac{5}{2}$ kernel with $N = 50$ training points (1000 seeds; shaded bands are 95\% confidence intervals).
        Each curve shows the selection probability of each boundary-distance bin after reweighting by shell volume; the dotted baseline is the geometric no-preference distribution.
        Lengthscales $\ell = 0.20, 0.30, 0.40, 0.55$ for $D = 2, 3, 4, 6$ are chosen as the smallest sweep value above the nearest-neighbor distance $N^{-1/D}$; \Cref{app:lengthscale_sweep} shows the bias is qualitatively robust across this range.
    }
    \Description{Four-panel line plot of selection probability versus boundary distance, one panel per dimension $D \in \{2, 3, 4, 6\}$. VM concentrates sharply at the boundary in every panel; NIPV and EPIG peak at interior shells, with NIPV's closer to the boundary and the two converging as $D$ grows.}
    \label{fig:selection_profiles}
\end{figure}

\Cref{fig:selection_profiles} shows the result.
The 2D pattern of \Cref{fig:surface_2d} persists across dimensions: VM stays at the boundary, while NIPV and EPIG settle onto interior shells whose locations depend on $\ell$ and $D$.
In low $D$, NIPV lies closer to the boundary; the shells converge as $D$ grows.
Mean acquisition-value profiles miss these patterns because the argmax is governed by the upper tail, not the mean (\Cref{app:argmax}).

The controlled diagnostic also matches sequential acquisition on representative test functions (\Cref{app:sequential}), confirming that the bias reflects kernel geometry rather than test-function properties.

\section{Discussion}
\label{sec:discussion}

Boundary variance inflation affects both BO and BED because it enters through the GP variance field.
In BO, this is not specific to UCB: any acquisition whose exploratory behavior is driven by posterior uncertainty can inherit the bias when exploration dominates.
We illustrate this with UCB, whose exploration term is explicit, and give a one-step BO example in \Cref{app:bo}.
Related effects can also arise in safe BO~\cite{sui2015safe}: inflated boundary uncertainty widens confidence bands and can make boundary regions harder to certify as safe.
In information-based design, the same effect can favor geometrically determined locations over the most informative ones.
Thus, the issue is not a peculiarity of one acquisition function, but a shared consequence of using a geometrically distorted variance field.

A Neumann (mirror-image) kernel mitigates the root cause by completing the truncated correlation neighborhood with reflected images (\Cref{app:neumann}).
As shown in \Cref{fig:neumann_correction}, it reduces VM's boundary concentration, removing $7\%$ of the excess over the geometric baseline at $D = 2$ and $28\%$ at $D = 6$.
The correction is partial: the image method changes the covariance geometry without adding independent observations beyond the domain.
More generally, reducing geometric nonuniformity in the variance field would decrease the shared source of bias across acquisition functions.

Our analysis isolates the variance-field mechanism under fixed hyperparameters; in full sequential workflows, hyperparameter learning and posterior-mean terms can attenuate or amplify it.
The selection-profile diagnostic provides a function-free way to check arbitrary acquisitions, kernels, and domain geometries for unintended spatial preferences.
Making these preferences measurable can help guide the design of acquisition functions and kernels with more reliable sampling behavior.

\section*{Code Availability}
Code for reproducing the experiments is available at
\url{https://github.com/mariabankestad/gp-boundary-bias}.

\section*{Author Contributions}
\label{app:authorship}
\textbf{Maria B\aa{}nkestad}: Conceptualization; Methodology; Software; Formal Analysis; Investigation; Visualization; Writing-Original Draft; Writing-Review \& Editing.
\textbf{Sanna Jarl}: Investigation; Writing-Review \& Editing.
\textbf{Jens Sj\"olund}: Conceptualization; Methodology; Writing-Review \& Editing.


\bibliographystyle{ACM-Reference-Format}
\bibliography{references}

@book{rasmussen2006gaussian,
  title={Gaussian Processes for Machine Learning},
  author={Rasmussen, Carl Edward and Williams, Christopher K. I.},
  year={2006},
  publisher={MIT Press},
  address={Cambridge, MA, USA}
}

@article{lindley1956eig,
  title={On a measure of the information provided by an experiment},
  author={Lindley, Dennis V.},
  journal={The Annals of Mathematical Statistics},
  volume={27},
  number={4},
  pages={986--1005},
  year={1956},
  publisher={Institute of Mathematical Statistics}
}

@article{mackay1992information,
  title={Information-Based Objective Functions for Active Data Selection},
  author={MacKay, David J. C.},
  journal={Neural Computation},
  volume={4},
  number={4},
  pages={590--604},
  year={1992},
  publisher={MIT Press},
  doi={10.1162/neco.1992.4.4.590}
}

@article{chaloner1995bayesian,
  title={Bayesian experimental design: A review},
  author={Chaloner, Kathryn and Verdinelli, Isabella},
  journal={Statistical Science},
  volume={10},
  number={3},
  pages={273--304},
  year={1995},
  publisher={Institute of Mathematical Statistics}
}

@inproceedings{bickfordsmith2023prediction,
  title={Prediction-Oriented {Bayesian} Active Learning},
  author={{Bickford Smith}, Freddie and Kirsch, Andreas and Farquhar, Sebastian and Gal, Yarin and Foster, Adam and Rainforth, Tom},
  booktitle={Proceedings of The 26th International Conference on Artificial Intelligence and Statistics},
  series={Proceedings of Machine Learning Research},
  volume={206},
  pages={7331--7348},
  year={2023},
  publisher={PMLR},
  address={Valencia, Spain}
}

@article{rainforth2024modern,
  title={Modern {Bayesian} experimental design},
  author={Rainforth, Tom and Foster, Adam and Ivanova, Desi R. and Bickford Smith, Freddie},
  journal={Statistical Science},
  volume={39},
  number={1},
  pages={100--114},
  year={2024},
  publisher={Institute of Mathematical Statistics}
}

@article{cohn1996active,
  title={Active learning with statistical models},
  author={Cohn, David A. and Ghahramani, Zoubin and Jordan, Michael I.},
  journal={Journal of Artificial Intelligence Research},
  volume={4},
  pages={129--145},
  year={1996}
}

@book{garnett2023bayesian,
  title={Bayesian Optimization},
  author={Garnett, Roman},
  year={2023},
  publisher={Cambridge University Press},
  address={Cambridge, UK}
}

@article{krause2008near,
  title={Near-Optimal Sensor Placements in {Gaussian} Processes: Theory, Efficient Algorithms and Empirical Studies},
  author={Krause, Andreas and Singh, Ajit and Guestrin, Carlos},
  journal={Journal of Machine Learning Research},
  volume={9},
  pages={235--284},
  year={2008}
}

@article{shahriari2016taking,
  title={Taking the human out of the loop: A review of {Bayesian} optimization},
  author={Shahriari, Bobak and Swersky, Kevin and Wang, Ziyu and Adams, Ryan P. and De Freitas, Nando},
  journal={Proceedings of the IEEE},
  volume={104},
  number={1},
  pages={148--175},
  year={2016},
  publisher={IEEE}
}

@book{vershynin2018high,
  title={High-Dimensional Probability: An Introduction with Applications in Data Science},
  author={Vershynin, Roman},
  year={2018},
  publisher={Cambridge University Press},
  address={Cambridge, UK}
}

@misc{surjanovic2013virtual,
  author={Surjanovic, Sonja and Bingham, Derek},
  title={Virtual Library of Simulation Experiments: Test Functions and Datasets},
  year={2013},
  url={https://www.sfu.ca/~ssurjano/optimization.html},
  note={Accessed: 2026-05-25}
}

@inproceedings{hvarfner2024vanilla,
  title={Vanilla {B}ayesian Optimization Performs Great in High Dimensions},
  author={Hvarfner, Carl and Hellsten, Erik Orm and Nardi, Luigi},
  booktitle={Proceedings of the 41st International Conference on Machine Learning},
  series={Proceedings of Machine Learning Research},
  volume={235},
  pages={20793--20817},
  year={2024},
  publisher={PMLR},
  address={Vienna, Austria}
}

@book{gramacy2020surrogates,
  title={Surrogates: {Gaussian} Process Modeling, Design, and Optimization for the Applied Sciences},
  author={Gramacy, Robert B.},
  year={2020},
  publisher={Chapman and Hall/CRC},
  address={Boca Raton, FL, USA}
}

@inproceedings{snoek2014input,
  title={Input Warping for {Bayesian} Optimization of Non-Stationary Functions},
  author={Snoek, Jasper and Swersky, Kevin and Zemel, Rich and Adams, Ryan},
  booktitle={Proceedings of the 31st International Conference on Machine Learning},
  series={Proceedings of Machine Learning Research},
  volume={32},
  pages={1674--1682},
  year={2014},
  publisher={PMLR},
  address={Beijing, China}
}

@inproceedings{srinivas2010gaussian,
  title={{Gaussian} process optimization in the bandit setting: No regret and experimental design},
  author={Srinivas, Niranjan and Krause, Andreas and Kakade, Sham M. and Seeger, Matthias},
  booktitle={International Conference on Machine Learning},
  pages={1015--1022},
  year={2010},
  publisher={Omnipress},
  address={Haifa, Israel}
}

@inproceedings{siivola2018boundary,
  title={Correcting boundary over-exploration deficiencies in {Bayesian} optimization with virtual derivative sign observations},
  author={Siivola, Eero and Vehtari, Aki and Vanhatalo, Jarno and Gonz{\'a}lez, Javier and Andersen, Michael Riis},
  booktitle={2018 IEEE 28th International Workshop on Machine Learning for Signal Processing (MLSP)},
  pages={1--6},
  year={2018},
  publisher={IEEE},
  address={Aalborg, Denmark},
  doi={10.1109/MLSP.2018.8516936}
}

@inproceedings{seo2000nipv,
  title={{Gaussian} process regression: Active data selection and test point rejection},
  author={Seo, Sambu and Wallat, Marko and Graepel, Thore and Obermayer, Klaus},
  booktitle={Proceedings of the IEEE-INNS-ENNS International Joint Conference on Neural Networks (IJCNN)},
  volume={3},
  pages={241--246},
  year={2000},
  publisher={IEEE},
  address={Como, Italy},
  doi={10.1109/IJCNN.2000.861310}
}

@article{gulian2022boundary,
  title={{Gaussian} process regression constrained by boundary value problems},
  author={Gulian, Mamikon and Frankel, Ari and Swiler, Laura},
  journal={Computer Methods in Applied Mechanics and Engineering},
  volume={388},
  pages={114117},
  year={2022},
  publisher={Elsevier},
  doi={10.1016/j.cma.2021.114117}
}

@inproceedings{solin2019boundary,
  title={Know Your Boundaries: Constraining {Gaussian} Processes by Variational Harmonic Features},
  author={Solin, Arno and Kok, Manon},
  booktitle={Proceedings of the Twenty-Second International Conference on Artificial Intelligence and Statistics},
  series={Proceedings of Machine Learning Research},
  volume={89},
  pages={2193--2202},
  year={2019},
  publisher={PMLR},
  address={Naha, Okinawa, Japan}
}

@book{cressie1993statistics,
  title={Statistics for Spatial Data},
  author={Cressie, Noel A. C.},
  edition={Revised},
  year={1993},
  publisher={Wiley},
  address={New York}
}

@inproceedings{knutsson1993normalized,
  title={Normalized and Differential Convolution: Methods for Interpolation and Filtering of Incomplete and Uncertain Data},
  author={Knutsson, Hans and Westin, Carl-Fredrik},
  booktitle={Proceedings of the IEEE Conference on Computer Vision and Pattern Recognition (CVPR)},
  pages={515--523},
  year={1993},
  publisher={IEEE},
  address={New York, NY, USA}
}

@inproceedings{sui2015safe,
  title={Safe Exploration for Optimization with {Gaussian} Processes},
  author={Sui, Yanan and Gotovos, Alkis and Burdick, Joel and Krause, Andreas},
  booktitle={Proceedings of the 32nd International Conference on Machine Learning},
  series={Proceedings of Machine Learning Research},
  volume={37},
  pages={997--1005},
  year={2015},
  publisher={PMLR},
  address={Lille, France}
}

\appendix
\crefalias{section}{appendix}

\section{Additional Theory and Mechanism}
\label{app:theory}

\subsection{Gaussian Process Posterior}
\label{app:gp}

Given observations $\mathcal{D} = \{X, \vy\}$, the GP predictive posterior at a test point $\vx_*$ is Gaussian with mean and variance
\begin{align}
\mu(\vx_*) &= m(\vx_*) + \vk_*^\top (K + \sigma_n^2 I)^{-1}(\vy - m(X)), \label{eq:gp_mean}\\
s^2(\vx_*) &= k_\theta(\vx_*, \vx_*) - \vk_*^\top (K + \sigma_n^2 I)^{-1} \vk_*,
\end{align}
where $K_{ij} = k_\theta(\vx_i, \vx_j)$ and $\vk_\vx = (k_\theta(\vx_1, \vx), \ldots, k_\theta(\vx_N, \vx))^\top$ is the vector of kernel evaluations between the training points and $\vx$.
The posterior covariance between two test points is
\begin{equation}
c(\vx, \vx') = k_\theta(\vx, \vx') - \vk_\vx^\top (K + \sigma_n^2 I)^{-1} \vk_{\vx'},
\end{equation}
with $s^2(\vx) = c(\vx, \vx)$ as the diagonal case.
The posterior mean $\mu(\vx)$ depends on the observed values $\vy$, but neither $s^2$ nor $c$ does.
This is immediate from the expressions above: both are determined entirely by $X$, $k_\theta$, and $\sigma_n^2$.

\subsection{One-Dimensional Variance Reduction Example}
\label{app:1d_example}

\begin{figure}[ht]
    \centering
    \includegraphics[width=0.7\textwidth]{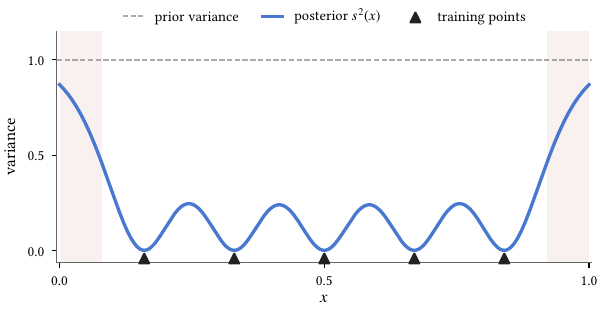}
    \caption{%
        Posterior variance $s^2(x)$ for a Mat\'ern-$\nicefrac{5}{2}$ kernel ($\ell = 0.12$, $\sigma_n^2 = 0.003$) given five evenly spaced training points on $[0, 1]$.
        Interior variance dips toward zero at each training point and rises moderately between them.
        Near the boundary ($x = 0$ and $x = 1$, shaded), the kernel's correlation neighborhood is truncated and the variance climbs back toward the prior level (dashed): the geometric artifact that drives the bias.\looseness-1
    }
    \Description{One-dimensional plot of GP posterior variance versus position on the unit interval. Variance dips toward zero at five interior training points and rises sharply toward the prior level near the two boundaries.}
    \label{fig:variance_1d}
\end{figure}

To make the truncated-neighborhood argument concrete, consider a Mat\'ern-$\nicefrac{5}{2}$ kernel in one dimension.
For a single training point $\vx_1$ at distance $h$ from $\vx_*$, the variance reduction is
\begin{equation*}
    \Delta s^2 = \frac{k_\theta(\vx_*, \vx_1)^2}{k_\theta(\vx_1, \vx_1) + \sigma_n^2},
\end{equation*}
which depends on $h/\ell$ but not on the position of $\vx_*$ relative to the boundary.
With multiple training points, the reduction is governed by the full covariance vector and kernel matrix, but the same intuition holds: fewer in-domain correlated observations are available near the boundary.
A boundary point, therefore, receives less variance reduction on average, producing the inflation visible in \Cref{fig:variance_1d}.

\subsection{Detailed Inheritance Analysis}
\label{app:inheritance}

\paragraph{Pointwise variance criteria: boundary concentration.}
For a GP with Gaussian likelihood, $\mathrm{EIG}(\vx) = \tfrac{1}{2}\log\bigl(1 + s^2(\vx)/\sigma_n^2\bigr)$ is a monotone transform of $s^2(\vx)$.
The boundary variance inflation therefore passes through directly: under the conditions of \Cref{sec:boundary_bias}, $\E_X[\mathrm{EIG}(\vx_\mathrm{bnd})] > \E_X[\mathrm{EIG}(\vx_\mathrm{int})]$, so the EIG argmax tends toward boundary locations.
The same holds for UCB's exploration term $\beta\sigma(\vx)$ and for pure variance maximization, since all three share the same variance-driven argmax.
Pointwise variance criteria therefore allocate observations according to kernel geometry rather than task-specific uncertainty.

\paragraph{NIPV: partial correction through global integration.}
NIPV evaluates each candidate by the extent to which it reduces variance across the entire domain.
A boundary candidate $\vx$ has high local variance $s^2(\vx)$, but its posterior covariances $c(\vx, \vx_j^*)$ with distant target points are truncated by the boundary, limiting the numerator $\sum_j c_j^2$.
An interior candidate has lower local variance but correlates symmetrically with a larger volume of the domain.
This trade-off displaces NIPV's argmax onto an interior shell just inside the boundary.
The shell's location depends on the lengthscale and dimension.

\paragraph{EPIG: a characteristic interior shell.}
EPIG averages pairwise mutual information terms $h(\rho_j^2) = -\frac{1}{2}\log(1-\rho_j^2)$ over target points.
The function $h$ is convex and steep near $\rho^2=1$.
This nonlinearity creates a qualitative difference from both EIG and NIPV:

\begin{enumerate}[label=(\roman*)]
    \item For a boundary candidate with $d_\partial(\vx) \ll \ell$, nearby targets on the interior side have high $\rho_j^2$ and contribute large EPIG terms.
    But targets on the far side of the domain have lower posterior correlation with $\vx$ than they would with an interior candidate, because the boundary truncates the propagation of correlations.
    The average over all targets is pulled down by these many poorly correlated distant targets.
    \item For an interior candidate with $d_\partial(\vx) \gg \ell$, the correlation field is symmetric and extends in all directions.
    Targets at moderate distance contribute a substantial $\rho_j^2$ from every direction, and the average is not truncated.
\end{enumerate}

The averaging acts as a spatial regularizer, penalizing boundary candidates.
But it does not produce spatially unbiased sampling.
The EPIG argmax concentrates at a characteristic interior shell at a boundary distance that depends on $\ell$ and $D$.
This shell arises because EPIG favors candidates with symmetric, far-reaching correlation neighborhoods, and there is a specific distance from the boundary where this symmetry is maximized relative to the competing effect of elevated prior variance.

\paragraph{NIPV-EPIG convergence as $D$ grows.}
NIPV and EPIG share the same cross-covariance information $c_j$ but aggregate it differently: NIPV linearly, EPIG through the saturating $h(\rho_j^2)$.
For $\rho_j^2 \ll 1$, the Taylor expansion gives $h(\rho_j^2) \approx \rho_j^2/2$, and substituting yields
\begin{equation*}
\mathrm{EPIG}(\vx) \;\approx\; \frac{1}{2M\,(s^2(\vx)+\sigma_n^2)} \sum_j \frac{c_j^2}{s^2(\vx_j^*)+\sigma_n^2},
\end{equation*}
which is NIPV with an additional per-target reweighting $1/(s^2(\vx_j^*)+\sigma_n^2)$.
If target variances are roughly uniform across the target set, the reweighting collapses to a constant and $\mathrm{EPIG} \propto \mathrm{NIPV}$ -- the two share an argmax.
Two effects push $\rho_j^2$ into this small-correlation regime as $D$ grows: with $N$ fixed and $\ell \sim N^{-1/D}$, sparse training keeps target variances close to the prior level and enlarges the EPIG denominator; and the product kernel $k(\vx, \vx_j^*) = \prod_d \kappa(x_d - x_{j,d}^*)$ attenuates sharply as soon as any coordinate distance exceeds $\ell$, so fewer candidate-target pairs reach high posterior correlation.
The convergence is gradual: profiles are noticeably similar from $D = 4$ and essentially coincide at $D = 6$ (right panel of \Cref{fig:selection_profiles}).

\section{Selection-Profile Diagnostic Details}
\label{app:diagnostic_details}

\subsection{Why the Argmax Matters, Not the Mean}
\label{app:argmax}

A natural first attempt at diagnosing spatial bias is to plot the mean acquisition value as a function of boundary distance: $\E[\alpha(\vx) \mid d_\partial(\vx) \in B_b]$.
This turns out to be misleading.
In our experiments, EPIG's mean profile is nearly flat across boundary-distance bins, suggesting a lack of spatial preference.
Yet the argmax of EPIG consistently falls in interior bins.
The reason is that sequential design selects the maximizer $\vx_\mathrm{next} = \argmax_{\vx \in \mathcal{X}_c} \alpha(\vx)$, so what determines sampling behavior is the upper tail of the acquisition distribution within each bin, not the conditional mean.
A bin with slightly higher variance in its acquisition values will more often produce the global maximum, even if its mean is the same as that of other bins.
This is why the selection profile records argmax locations rather than mean acquisition values.

A related concern is that, in practice, BO uses approximate acquisition optimization rather than the exact grid argmax, which could, in principle, dilute the bias by missing the exact maximizer.
Two observations argue against this.
First, the bias concerns \emph{where} the high-acquisition regions sit: any optimizer that climbs the acquisition surface is pulled into the boundary-inflated regions, regardless of whether it recovers the exact argmax.
Second, the sequential validation of \Cref{fig:combined_bias} already uses MAP-fitted GPs with realistic acquisition optimization on real test functions, and the spatial bias persists: it is not an artifact of exact-argmax recovery on a candidate grid.

\subsection{Controlled Diagnostic Setup}
\label{app:experimental}

We fix the dimension $D$, training set size $N$, and kernel hyperparameters $(\ell, \sigma_n^2)$, then repeat the following $S = 1000$ times.
Draw $N$ training points uniformly on $[0,1]^D$ and compute the GP posterior under a product Mat\'ern-$\nicefrac{5}{2}$ kernel.
Evaluate each acquisition function on a candidate set $\mathcal{X}_c$ and record which boundary-distance bin contains the argmax.
The selection probability for bin $b$ is
\begin{equation}
    \hat{p}_b \;=\; \frac{1}{S}\sum_{s=1}^{S} \mathbf{1}\!\bigl[d_\partial(\vx_\mathrm{next}^{(s)}) \in B_b\bigr].
    \label{eq:selection_prob}
\end{equation}

For the headline figures, the lengthscale at each $D$ is chosen as the smallest value in the sweep grid of \Cref{app:lengthscale_sweep} above the typical nearest-neighbor distance $N^{-1/D}$ of $N = 50$ uniform training points; this places the kernel range $2\ell$ comfortably above the inter-point spacing, the regime in which the GP meaningfully interpolates between training points.
The specific settings are $N = 50$ throughout, with $\ell = 0.20$ for $D = 2$, $\ell = 0.30$ for $D = 3$, $\ell = 0.40$ for $D = 4$, and $\ell = 0.55$ for $D = 6$; observation noise $\sigma_n^2 = 0.003$.
For the 2D surface figure (\Cref{fig:surface_2d}), we use an isotropic Mat\'ern-$\nicefrac{5}{2}$ kernel with $\ell = 0.2$, $N = 15$, $\sigma_n^2 = 0.003$, evaluated on a $200 \times 200$ grid with 2048 Sobol integration points; the argmax density aggregates 10\,000 random training sets on a $128 \times 128$ candidate grid.

\subsection{Stratification and Volume Reweighting}
\label{app:reweighting}

The unit hypercube $[0,1]^D$ has far more volume near its boundary than in its interior~\cite{vershynin2018high}.
Under uniform sampling, the density of $d_\partial$ is $f(t) = 2D(1-2t)^{D-1}$, which is sharply concentrated near $t = 0$ for moderate~$D$.
If candidates are drawn uniformly, the selection profile mixes two effects: the acquisition function's preference and the cube's geometric volume.

To separate them, we build a stratified candidate set in which each boundary-distance bin contains the same number of candidates.
This gives each shell equal opportunity to contain the argmax and reveals the acquisition's intrinsic preference over $d_\partial$.
For plots that represent uniform candidate availability in the cube, we then reweight the stratified bin probabilities by the shell volumes
\[
w_b \propto (1 - 2a_b)^D - (1 - 2b_b)^D,
\]
where $a_b$ and $b_b$ are the bin edges.
The dashed baseline is the corresponding no-preference distribution: the distribution obtained by an acquisition function that is equally likely to select any uniformly available candidate.

\section{Robustness and Sequential Validation}
\label{app:validation}

\subsection{Sequential Validation}
\label{app:sequential}

To confirm that the controlled diagnostic predicts real sequential behavior, we run acquisition on three test functions: Peaks (MATLAB) ($D = 2$, $N_0 = 35$), Smooth-box, a product of one-dimensional smooth step functions ($D = 3$, $N_0 = 40$), and Hartmann~\cite{surjanovic2013virtual} ($D = 4$, $N_0 = 50$).
Starting from $N_0$ initial points, we fit a GP with MAP hyperparameters under a dimension-scaled log-normal lengthscale prior~\cite{hvarfner2024vanilla} and select one point via each acquisition function.
Using 1000 independent seeds, we histogram $d_\partial(\vx_\mathrm{next})$ and compare it to the controlled diagnostic.

\begin{figure}[t]
    \centering
    \includegraphics[width=0.8\textwidth]{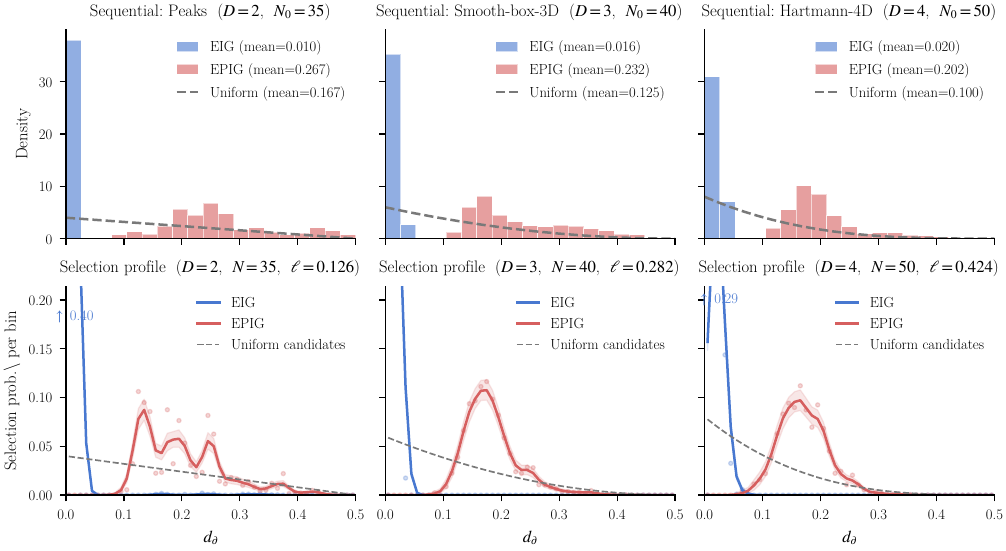}
    \caption{%
        \textbf{Top row:} Histograms of boundary distance $d_\partial$ of points selected in sequential acquisition on test functions (Peaks $D{=}2$, Smooth-box $D{=}3$, Hartmann $D{=}4$; 1000 seeds each).
        The dashed line is the geometric density of $d_\partial$ under uniform sampling from $[0,1]^D$.
        \textbf{Bottom row:} Selection profiles from the controlled diagnostic with hyperparameters matched to the sequential experiments (1000 seeds).
        Faint dots show raw bin probabilities; solid lines are smoothed; shaded bands are 95\% bootstrap confidence intervals.
        The dashed baseline is the geometric uniform-candidate distribution.
        VM concentrates at the domain boundary; EPIG concentrates at a characteristic interior shell.
    }
    \Description{Two-row figure comparing sequential acquisition (top) with the controlled diagnostic (bottom) on three test functions across $D = 2, 3, 4$. Both rows show VM concentrated at the boundary and EPIG concentrated at an interior shell, in close agreement.}
    \label{fig:combined_bias}
\end{figure}

\Cref{fig:combined_bias} validates the controlled diagnostic against sequential acquisition on three test functions with fitted GP hyperparameters.
The top row shows histograms of $d_\partial$ for the first point selected after the initial design, aggregated over 1000 independent seeds.
The bottom row shows the controlled diagnostic at the median hyperparameters fitted in the corresponding sequential experiments.
The close agreement between the two rows confirms that the spatial biases visible in the diagnostic are not artifacts of the simplified setup; they persist in full sequential acquisition with MAP-fitted hyperparameters on real test functions.
Note that the figure labels use ``EIG'' rather than ``VM'' because EIG is a monotone transform of the posterior variance; thus, EIG and VM share the same argmax, and their selection profiles are identical.

\subsection{Lengthscale Robustness}
\label{app:lengthscale_sweep}

\begin{figure}[t]
    \centering
    \includegraphics[width=0.9\textwidth]{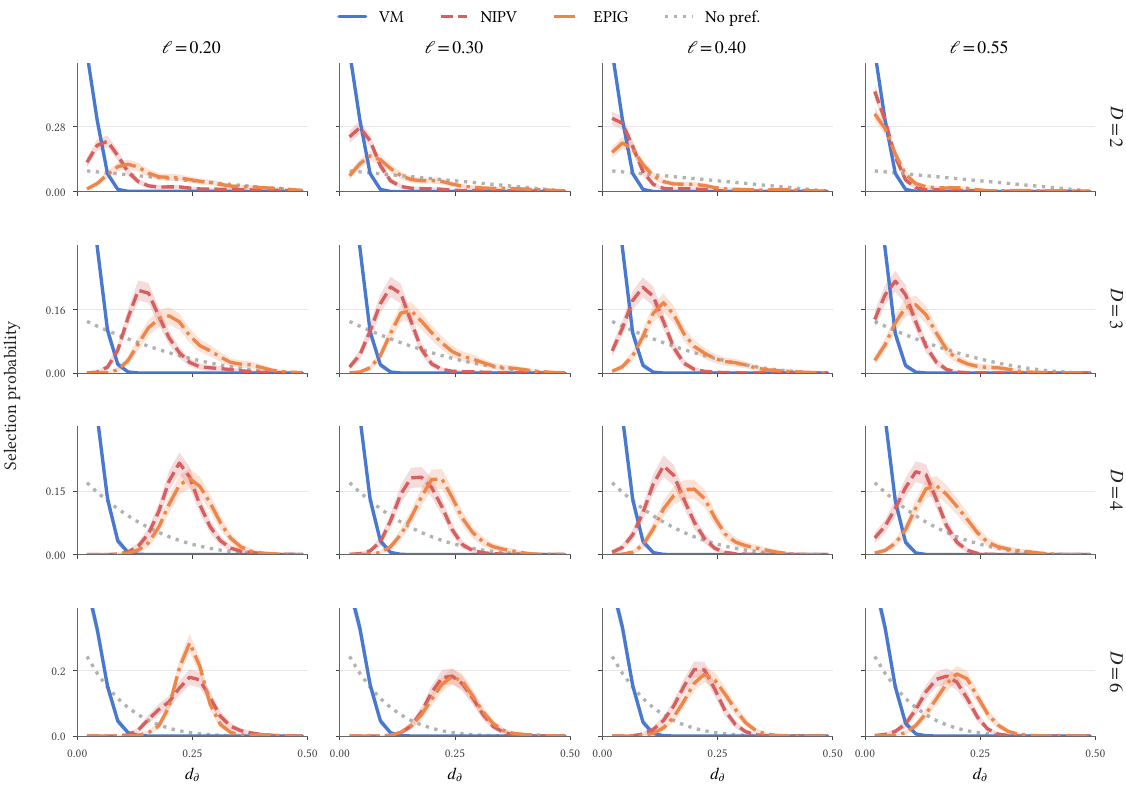}
    \caption{%
        Selection profiles across a lengthscale sweep at $N = 50$ training points (1000 seeds per setting).
        Rows: dimension $D \in \{2, 3, 4, 6\}$; columns: lengthscale $\ell \in \{0.20, 0.30, 0.40, 0.55\}$.
        Each panel shows the volume-reweighted acquisition-argmax distribution for variance maximization (blue), negative integrated posterior variance (red), and expected predictive information gain (orange), against the geometric no-preference baseline (dotted gray).
        VM concentrates at the boundary in every $(D, \ell)$ setting: the pattern persists across the entire lengthscale range.
        NIPV and EPIG sit at the interior in most settings; their boundary fraction grows at large $\ell$ in low $D$, where the kernel range becomes a substantial fraction of the domain, and the boundary truncation begins to dominate the global integration.\looseness-1
    }
    \Description{Four-by-four grid of selection-probability curves across dimensions (rows: $D = 2, 3, 4, 6$) and lengthscales (columns: $\ell = 0.20, 0.30, 0.40, 0.55$). VM concentrates at the boundary in every panel; NIPV and EPIG sit at the interior in most panels, drifting toward the boundary only at the largest $\ell$ in low $D$.}
    \label{fig:sweep_profiles}
\end{figure}

\Cref{fig:sweep_profiles} extends the controlled diagnostic across a sweep of lengthscales $\ell \in \{0.20, 0.30, 0.40, 0.55\}$ at $N = 50$ training points, for $D \in \{2, 3, 4, 6\}$ (1000 random training sets per setting).
The figure isolates the dependence of each acquisition's spatial preference on the kernel lengthscale.

\begin{figure}[t]
    \centering
    \includegraphics[width=0.9\textwidth]{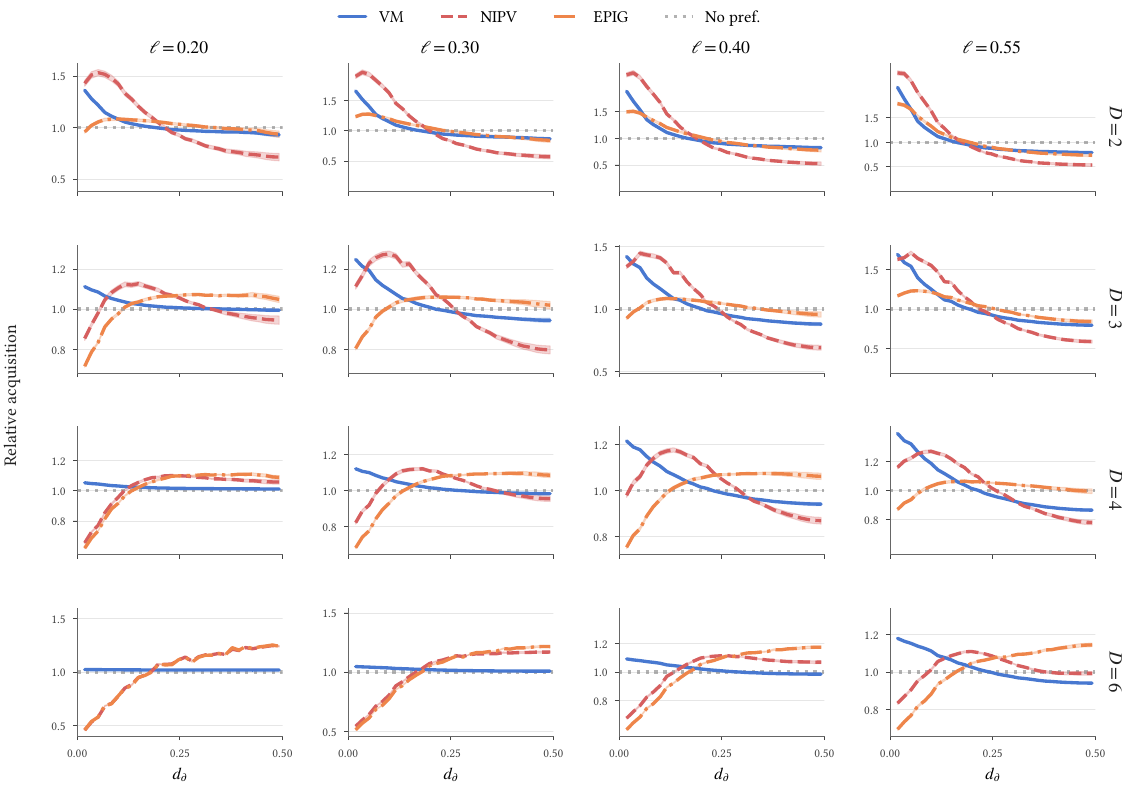}
    \caption{%
        Normalized mean acquisition-value profiles across the same sweep as \Cref{fig:sweep_profiles} (1000 seeds per setting).
        Curves are per-seed normalized so the integral over $d_\partial$ equals $0.5$, so a perfectly flat profile sits at $1.0$ (the dotted reference).
        Shaded bands are 95\% confidence intervals on the across-seed mean.
        Y-axes are scaled per panel and centered on $1.0$ so the within-panel deviation structure is visible at each panel's natural scale.
        Mean profiles remain narrowly clustered around $1$ in every panel, while the corresponding selection profiles (\Cref{fig:sweep_profiles}) show sharp boundary or interior preferences.
    }
    \Description{Four-by-four grid of normalized mean acquisition value versus boundary distance, across dimensions and lengthscales. Curves remain narrowly clustered around one in every panel, in contrast to the sharp spatial preferences visible in the corresponding selection-profile figure.}
    \label{fig:acq_sweep}
\end{figure}

\Cref{fig:acq_sweep} shows normalized mean acquisition-value profiles across the same sweep: rows are $D \in \{2, 3, 4, 6\}$, columns are $\ell \in \{0.20, 0.30, 0.40, 0.55\}$.
In every panel, the normalized mean acquisition value stays close to one across $d_\partial$ (the dotted reference), even where the corresponding selection profile (\Cref{fig:sweep_profiles}) shows a sharp spatial preference, reinforcing the argmax-versus-mean point of \Cref{app:argmax} across the full robustness sweep.

\section{One-Step BO Placement}
\label{app:bo}

The selection-profile diagnostic isolates the observation-independent family (VM, NIPV, EPIG), whose spatial preference, under fixed hyperparameters, is determined by the kernel geometry and the training locations alone (\Cref{rem:y_independence}).
Full BO acquisitions (EI, UCB, PI) add a posterior-mean component, and the boundary inflation enters only through the variance term.
Whether the bias survives in practice depends on the regime: at small $N$ the mean is uninformative, and exploration dominates; as data accumulates, the mean takes over.

We extend the diagnostic to a one-step BO setting on the 2D Gramacy--Lee function~\cite{surjanovic2013virtual} $f(x_1, x_2) = x_1 \exp(-x_1^2 - x_2^2)$ (mapped from $[0,1]^2$ to $[-2,6]^2$), which has localized structure in the lower-left quadrant and is essentially flat elsewhere: a regime where the posterior mean is informative only in a small region.
For each $N_0 \in \{5, 10, 25\}$ and $10\,000$ Sobol initial designs, we condition a GP with an isotropic Mat\'ern-$\nicefrac{5}{2}$ kernel (fixed $\ell = 0.25$, fixed noise at $5\%$ of $\mathrm{std}(f)$) and record the argmax of two acquisitions on a $100 \times 100$ candidate grid: pure variance $\mathrm{VM}(\vx) = s^2(\vx)$ as the mean-free reference, and $\mathrm{UCB}(\vx) = \mu(\vx) + \kappa\,s(\vx)$ with $\kappa = 2$~\cite{srinivas2010gaussian}.
Sobol initial designs control for the interior-gap confound: with a well-spread initial set, there are no large random interior gaps, so a boundary selection reflects the inflation rather than a sparsity accident.

\begin{figure}[t]
    \centering
    \includegraphics[width=\textwidth]{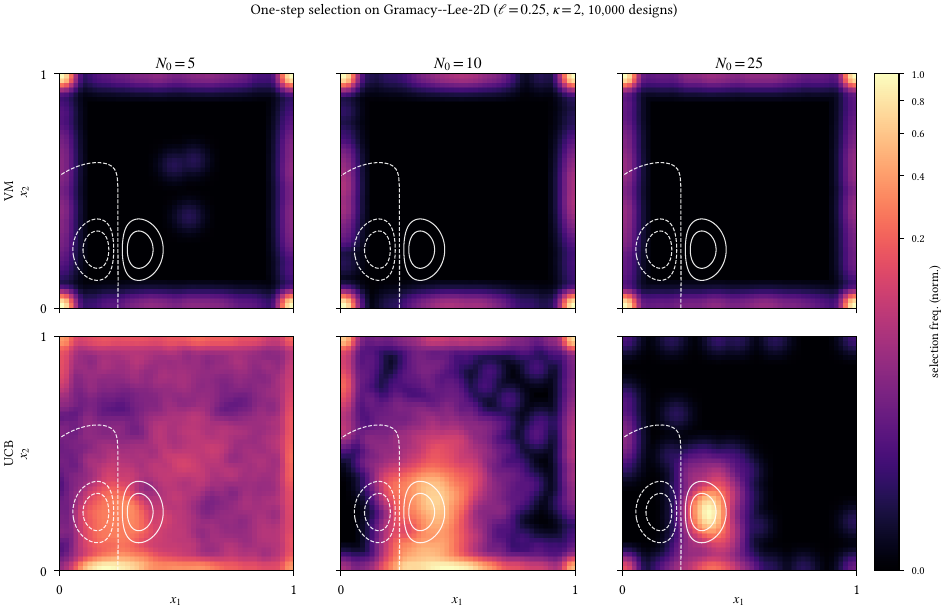}
    \caption{%
        One-step BO placement on the 2D Gramacy--Lee function with an isotropic Mat\'ern-$\nicefrac{5}{2}$ kernel ($\ell = 0.25$) and Sobol initial designs of size $N_0$.
        \textbf{Top row} (VM, posterior variance only): selection density locked at the corners across all $N_0$, independent of the data.
        \textbf{Bottom row} (UCB, $\mu + 2\sigma$): at small $N_0$ the mean is uninformative, and the boundary-inflated variance dominates, with $\sim 48\%$ of selections near the boundary at $N_0 = 5$; as $N_0$ grows, the mean takes over and selection concentrates on the function's interior structure.
        Per-panel-normalized densities over $10\,000$ random initial designs, power-law color scale.
        Function contours shown for reference (solid = positive levels, dashed = negative).
    }
    \Description{Two-by-three grid of 2D selection-density heatmaps overlaid with function contours. The top row (VM) shows density locked at the four corners for all three initial design sizes. Bottom row (UCB) shows density at the corners at small $N_0$, transitioning to the function's interior structure at larger $N_0$.}
    \label{fig:bo_2d}
\end{figure}

\Cref{fig:bo_2d} shows the result.
VM places all selections within $d_\partial < 0.05$ of the boundary at every $N_0$, the data-blind corner concentration isolated by the function-free diagnostic, now confirmed in the presence of training data and a fitted GP.
UCB transitions with $N_0$: $48\%$ of selections are near-boundary at $N_0 = 5$, $40\%$ at $N_0 = 10$, and $15\%$ at $N_0 = 25$, compared with a geometric baseline of $\sim 16\%$ under uniform sampling.
The boundary pull is strongest at small $N_0$, where the posterior mean is weakly informative and the boundary-inflated variance dominates.
At $N_0 = 25$, UCB is close to the geometric baseline, indicating that the mean component can eventually overcome the variance artifact when enough data are available.

\section{Neumann Kernel Correction}
\label{app:neumann}

The root cause of boundary variance inflation is the truncation of the kernel's correlation neighborhood at the domain boundary.
A Neumann, or mirror-image, kernel addresses this by extending the covariance through reflected images, producing an even extension that satisfies Neumann (zero normal derivative) boundary conditions on $[0,1]^D$.

For a stationary one-dimensional base kernel $\kappa(r)$, the Neumann kernel is approximated by the finite image sum
\begin{equation}
    k_N(x, x') = \sum_{n=-L}^{L} \bigl[\kappa(x - x' + 2n) + \kappa(x + x' + 2n)\bigr],
    \label{eq:neumann_1d}
\end{equation}
where $L$ controls the number of image layers.
The first term accounts for periodic translations and the second for reflections about the boundary.
In all experiments, $L$ is chosen large enough that additional image layers have negligible contribution at the lengthscales considered.
Together, the image terms make the effective correlation neighborhood more symmetric near $x=0$ and $x=1$.

For a product kernel on $[0,1]^D$, we apply this construction dimension-wise and normalize to unit diagonal:
\begin{equation}
    \tilde{k}_N(\vx, \vx') = \frac{k_N(\vx, \vx')}{\sqrt{k_N(\vx, \vx)\, k_N(\vx', \vx')}}.
    \label{eq:neumann_normalized}
\end{equation}
The normalization is important: without it, the unnormalized kernel has a position-dependent diagonal that would inflate the prior variance near boundaries, counteracting the correction.
With normalization, the correction arises from off-diagonal correlations: a boundary point gains correlations with training points on the ``reflected'' side, increasing the effective number of contributing neighbors.

This is not a claim that the unknown function satisfies Neumann boundary conditions.
Rather, the mirror-image construction corrects the geometric artifact in the posterior variance that causes boundary oversampling.

\begin{figure}[b]
    \centering
    \includegraphics[width=0.85\textwidth]{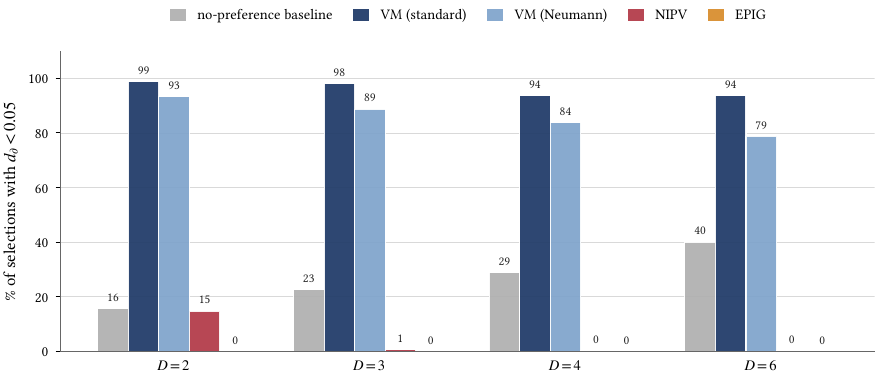}
    \caption{%
        Selection-profile response to the Neumann correction at the matched controlled-diagnostic settings of \Cref{fig:selection_profiles} ($N = 50$).
        Fraction of acquisition argmax within $d_\partial < 0.05$ under: the geometric no-preference baseline (gray); VM with the standard product Mat\'ern kernel (blue); VM with the Neumann kernel (green); and NIPV (red) and EPIG (orange) with the standard kernel for context.
        VM has a strong boundary bias, and Neumann shifts it toward the baseline, with the reduction increasing as $D$ increases.
        NIPV is intermediate at $D=2$ and self-corrects at higher $D$; EPIG sits below the baseline throughout. Both penalize boundary candidates via their respective integration structures, so the kernel modification is most relevant to VM.
    }
    \Description{Bar chart with four groups of five bars showing percent near-boundary selections at dimensions 2, 3, 4, and 6 for: a no-preference geometric baseline, VM with the standard kernel, VM with the Neumann kernel, NIPV, and EPIG. VM stays at 94-99\% across all dimensions; the Neumann variant shifts it downward by 5-15 percentage points.}
    \label{fig:neumann_correction}
\end{figure}

\Cref{fig:neumann_correction} shows the diagnostic's response to swapping the standard product Mat\'ern kernel for the Neumann kernel, at the same controlled-diagnostic settings as \Cref{fig:selection_profiles} ($N = 50$, $\ell$ per $D$ as listed).
Across $D \in \{2, 3, 4, 6\}$, the Neumann kernel reduces the fraction of VM selections within $d_\partial < 0.05$ of the boundary by 6 to 15 percentage points, with the reduction growing with $D$.
Expressed as a fraction of the excess over the geometric no-preference baseline, the correction removes $7\%$ at $D = 2$, $12\%$ at $D = 3$, $15\%$ at $D = 4$, and $28\%$ at $D = 6$, consistent with the $2^D$ corner amplification of the product construction.
The baseline shown is the geometric $P(d_\partial < 0.05 \mid d_\partial \in [0.01, 0.5])$ to match the candidate grid's range; selections use the volume-reweighted argmax of \Cref{fig:selection_profiles}.
For context, \Cref{fig:neumann_correction} also shows NIPV and EPIG at the same settings, with the standard kernel.
EPIG sits at $\sim 0\%$ in every dimension, well below the baseline: its averaging over targets already penalizes boundary candidates.
NIPV is intermediate at $D = 2$ ($15\%$, just below the $16\%$ baseline) and drops to $\sim 0\%$ for $D \in \{3, 4, 6\}$, matching the ``partial correction through global integration'' picture from \Cref{app:inheritance}.
The kernel-level Neumann modification is therefore most consequential for VM, the acquisition whose argmax most directly inherits the boundary inflation.

The correction is partial in every dimension.
Reflection modifies the covariance geometry but does not introduce independent observations outside the domain, so residual boundary preference persists.
Fully equalizing the variance field may require nonstationary kernels that explicitly encode the boundary, or input warping~\cite{snoek2014input} that maps boundary regions inward; both directions are left for follow-up work.
Here, the Neumann result serves primarily as a causal test of the diagnosis: modifying the truncated correlation neighborhood shifts the selection profile in the predicted direction, confirming that the bias is geometric in origin.

\end{document}